\documentclass[runningheads]{llncs}
\usepackage[T1]{fontenc}
\usepackage{graphicx}

\usepackage{amsmath,amssymb,amsfonts}
\usepackage{algorithmic}
\usepackage{textcomp}
\usepackage{xcolor}
\usepackage{multirow}
\usepackage{makecell}
\usepackage{multicol}
\usepackage{algorithm}
\usepackage{subfigure}
\begin{document}
\title{LLM Compression with Jointly Optimizing Architectural and Quantization choices}
%
%
\author{Hoang-Loc La\inst{1} \and
Truong-Thanh Le\inst{1,2} \and
Amir Taherkordi\inst{2} \and 
Phuong Ha Hoai\inst{1}}
%
\institute{UiT The Arctic University of Norway \and
University of Oslo, Norway \\
\email{\{hoang.l.la,phuong.ha.hoai\}@uit.no \\\{truongl,amirhost\}@ifi.uio.no}}

\maketitle              
\vspace{-0.5cm}
\begin{abstract}
Deploying large language models (LLMs) is challenging due to their significant memory and computational requirements. While some methods address this by developing small or tiny language models from scratch, these approaches demand extensive GPU training. Compressing pre-trained LLMs for edge devices offers a compelling alternative. Beyond pruning and quantization, Neural Architecture Search (NAS) enables effective compression, yet prior NAS approaches often limit the search space and decouple architecture from quantization. We introduce a differentiable NAS framework that explores the entire space and jointly optimizes architectural configurations alongside mixed-precision quantization for linear layers of LLMs. Experiments demonstrate superior accuracy-latency trade-offs: our models achieve up to $1.4×\times$ faster inference than sequential NAS-then-quantization baselines at comparable accuracy, or up to 6\% higher average accuracy across seven reasoning tasks at equivalent latency.

\keywords{LLM Compression \and Efficient LLM}
\end{abstract}

\vspace{-0.5cm}
\section{Introduction}

\par In recent years, Large Language Models (LLMs) have gained widespread attention, but their high computational and memory demands make deployment difficult on resource-constrained devices like laptops and smartphones. Rising privacy concerns with cloud-based LLMs have fueled interest in on-device inference, yet memory requirements remain a key barrier.
\par Two main approaches address the deployment challenges of LLMs: developing novel lightweight language models and compressing existing pre-trained LLMs. The first method involves training small language models from scratch, such as TinyLlama \cite{zhang2024tinyllama} with 1 billion parameters, which requires 90 days on 16 A100-40GB GPUs. Similarly, training Phi-2 \cite{javaheripi2023phi} with 2.7 billion parameters takes 14 days on 96 A100 GPUs.
\par The second approach leverages pre-trained LLMs, avoiding training from scratch and significantly reducing training time. Alongside techniques like structural pruning \cite{ashkboos2024slicegpt} and quantization \cite{frantar2022gptq}, neural architecture search (NAS) has emerged as a key method for LLM compression. However, current NAS applications for LLM compression face challenges, such as requiring resource-intensive supernet training \cite{caillamaflex,cai2024flextron} or updating only a small subset of candidate sub-networks during supernet training \cite{sukthanker2024large,munoz2024lonas}.

\par In contrast to prior works that explore only a limited portion of the search space~\cite{sukthanker2024large,munoz2024lonas,cai2024flextron}, Our proposed NAS framework optimizes over the full defined discrete search space via a relaxation. By directly optimizing architecture parameters under given constraints— without pre-selection as in \cite{sukthanker2024large}, our method explores a broader search space.
\par Furthermore, unlike conventional approaches that apply uniform quantization post-pruning, our method jointly optimizes both architectural configurations and layer-specific non-uniform quantization policies. This enables the discovery of optimal pruning-quantization combinations, achieving substantially lower memory usage and superior accuracy compared to state-of-the-art methods. Our main contributions are as follows.
\begin{itemize}
\item We introduce a novel differential weight-entanglement supernet design together with a constrained differential optimization method for efficient compression of pre-trained LLMs. Our approach achieves superior accuracy and lower latency compared to the state-of-the-art methods.
\item We present the first unified NAS framework that simultaneously optimizes model architecture and layer-wise quantization precision for LLM, addressing the longstanding limitation of treating pruning and quantization as separate steps. Models found by our joint approach deliver up to $1.4 \times$ faster inference than those produced by sequential NAS-then-quantization pipelines.

\item   We develop a novel vectorized implementation that significantly accelerates training of weight-entanglement supernets for LLMs, reducing training time by up to 4 $\times$ compared to the original approach \cite{sukthanker2024weight}.
\end{itemize}

    
    
\vspace{-0.3cm}
\section{Related Work and Our Advancements}
\vspace{-0.3cm}

\label{Sect:related_work}
\subsection{Weight-Entanglement NAS}
\par Sukthanker et al.~\cite{sukthanker2024weight} introduce a weight superposition technique in TangleNAS, which consolidates all possible weight matrix configurations into a single weighted representation, with each configuration assigned a learned importance scalar. This enables simultaneous training and architecture search in a single stage, offering the potential for reduced search costs. However, applying TangleNAS directly to large language models (LLMs) presents significant challenges.
\begin{itemize}
    \item The mixed-operation approach from TangleNAS \cite{sukthanker2024weight} is not suitable for compressing the depth dimension of LLMs. It restricts depth reduction to dropping only the final consecutive blocks, whereas the importance of individual blocks in a pre-trained foundation model varies substantially. Removing a more critical block can lead to substantial accuracy degradation.
    \item The weight-entanglement mechanism in TangleNAS was not optimized for GPU efficiency, making it impractical for training supernet at the scale required for LLMs (see Section \ref{sec:software} for further discussion).
\end{itemize}

\par \textbf{Our contribution}: We propose an efficient weight-entanglement-based supernet design specifically tailored for compressing large language models. Our framework significantly expands the search space by incorporating diverse quantization precision options across layers. To address the first limitation of TangleNAS, we propose importance-aware depth pruning that enables more flexible and effective depth compression (detailed in Section \ref{subsubsect:depth_dimension}). To overcome the second limitation, we develop a software-level optimization that substantially accelerates supernet training, making it feasible to compress LLMs using only a single NVIDIA A100 80GB GPU within practical time budgets. 

\vspace*{-3mm}
\subsection{Neural Architecture Search techniques for LLM compression}

A notable feature of transformers is permutation equivariance, which allows reordering embedding features, MLP intermediate features, and attention heads without significantly impacting model accuracy \cite{xu2024permutation}. Leveraging this, a common preprocessing step involves ranking the components of a pre-trained LLM by importance \cite{sukthanker2024large}. When selecting a sub-network, we can then simply choose the first neurons or heads from the supernet. This preprocessing method is widely used in the related work \cite{cai2024flextron,caillamaflex,sukthanker2024large} and is also adopted in our study.



\par LoNAS \cite{munoz2024lonas} and \textit{subnet-selection} \cite{sukthanker2024large} adopted a two-stage neural architecture search (NAS) approach to identify optimal sub-architectures in a pre-trained large language model (LLM). In the first stage, they train a supernet using LoRA\cite{hu2022lora}. In the second stage, they employ a multi-objective search to find sub-architectures that optimize accuracy and performance metrics, such as latency and energy efficiency, with the supernet serving as an accuracy estimator for sub-networks.

\par Unlike LoNAS \cite{munoz2024lonas}, \textit{subnet-selection} method \cite{sukthanker2024large} incorporates a pre-selection step before the NAS process. They rank the model’s features and blocks by importance and select the top neurons, heads, or blocks when sampling sub-networks. They observed that random sampling, as used in LoNAS, introduces bias in the search space, where smaller sub-architectures are updated more frequently than larger ones, complicating supernet training. To address this and optimize the search space, they introduced a grid-based sampling approach, dividing the search space into $K$ partitions and selecting the best candidate from each. During supernet fine-tuning, they randomly select $k \ll K$ sub-networks and use knowledge distillation to train these sub-networks alongside the largest (original) network. However, this heuristic approach can introduce a strong bias based on the initial selection criteria, potentially resulting in suboptimal architectures from the beginning. 

\par \textbf{Our contribution}: Previous techniques focus primarily on the often-updated subnets while disregarding the bulk of those that are rarely or never updated, which may lead to missing the genuine optimal solution. In this work, we present a novel differential neural architecture search (NAS) method designed for compressing pre-trained large language models (LLMs). By leveraging weight entanglement style supernet, our differential supernet does not require randomly sampling architectures from search space, and thus avoids the skewness of architecture distribution as in LoNAS. Meanwhile, our method does not rely on any heuristic to pre-select architectures as in \textit{subnet-selection}. In contrast, our method explores all conceivable candidate subnets during fine-tuning, progressively converging toward the optimal structure. Additionally, unlike earlier methods that lack support for quantization or require to apply compression and quantization techniques sequentially, our approach simultaneously optimizes architectural parameters and quantization precision across layers.
\vspace{-0.3cm}
\section{Method}

\label{Sect:Method}
\subsection{Constrained Differential NAS}
\subsubsection{Problem Formulation}
\par  We approach the compression of large language models (LLMs) as a constrained optimization problem. Our search space $S$ is parameterized by $\zeta \in S$, governing the architectural structure in a fully differentiable manner similar as in \cite{sukthanker2024weight}. A sampled candidate network is drawn as $\hat{\zeta} \sim P_{\zeta}(S)$, where $P_{\zeta}(S)$ is a probability distribution over search space $S$, parameterized by $\zeta$. $\mathcal{L}_\text{train}$ and $\mathcal{L}_\text{val}$ denote the training and validation losses, respectively. Consequently, the LLM compression task can be formulated as a bi-level constrained optimization problem as described below. 

\vspace{-0.3cm}
\begin{subequations}
\vspace{-0.3cm}
\begin{align} 
\zeta^* = \quad & \min_{\zeta}  \mathcal{L}_\text{val}(w^*, \hat{\zeta}) 
\label{eq:1}  \\
\text{s.t.} \quad & w^* = \arg\min_{w} \mathcal{L}_\text{train}(w, \hat{\zeta}), \notag \\
\quad &  B_\text{min} < F_\text{params}(\zeta_{\text{discrete}}) < B_\text{max}  \label{eq:1_b}  
\end{align}
\vspace{-0.3cm}
\end{subequations}
\vspace{-0.3cm}
\par Denote $w$ as weight of the pre-trained model. Let $F_\text{params}(\hat{\zeta})$ denotes the parameter count of the optimal neural architecture derived from the discretization step within the NAS procedure. This discretization step entails applying the $\arg\max$ function to the architectural parameters $\zeta$ in order to select the definitive architecture, yielding $\zeta_{\text{discrete}} = \arg\max(\zeta)$. The involvement of the $\arg\max$ operation renders $F_\text{params}$ non-differentiable. A simple method for approximating $F_\text{params}(\zeta_{\text{discrete}})$ involves relaxing the hard constraint  by calculating the expected value $\mathbf{E}_{\hat{\zeta} \sim P_{\zeta}(S)}[F_\text{params}(\hat{\zeta})]$.

\vspace*{-2mm}
\subsubsection{Constrained Optimization}
\label{subsect:constrainted_op}
\par We can transform the constrained optimization into an unconstrained one by adding a regularization term for the constraint \ref{eq:1_b}. Particularly, the constraint \ref{eq:1_b} can be formalized with a pair of ReLU functions as follows:
\vspace*{-1mm}
\begin{align}
F_{\text{constraint}} = \; & ReLU(\mathbf{E}_{\hat{\zeta} \sim P_{\zeta}(S)}[F_\text{params}(\hat{\zeta})] - B_\text{max})  \notag \\
\; &  +ReLU(B_\text{min} - \mathbf{E}_{\hat{\zeta} \sim P_{\zeta}(S)}[F_\text{params}(\hat{\zeta})])
\end{align} 
\vspace*{-1mm}
The loss terms in Equation \ref{eq:1} can be reformulated as:
\vspace*{0mm}
\begin{align} \label{eq:2}
\mathcal{L}_\text{train}(w, \hat{\zeta}) = \; & \mathbf{E}_{\hat{\zeta} \sim P_{\zeta}(S)} [\mathcal{L}_{\text{CE}}(w, \hat{\zeta})]  \\ 
\mathcal{L}_\text{val}(w^*, \zeta) = \; & \mathbf{E}_{\hat{\zeta} \sim P_{\zeta}(S)} [\mathcal{L}_{\text{CE}}(w^*, \hat{\zeta}) + \eta F_\text{latency}(\hat{\zeta})] \notag \\
\; & + \lambda F_{\text{constraint}} \notag
\end{align} 
\vspace*{-5mm}

\par where $L_{\text{CE}}$ is the cross-entropy loss and $F_{\text{latency}}(\hat{\zeta})$ is the expected inference latency of the sampled architecture. This latency term is computed as a probability-weighted average of per-choice latencies, which are retrieved from a pre-calculated lookup table. We denote $\eta$ is a hyperparameter defining the trade-off between validation loss $L_{\text{CE}}$ and inference latency $F_{\text{latency}}$. $\lambda$ is a hyperparameter to control strength of the regularization term $F_{\text{constraint}}$. It is note-worthy that $F_{\text{latency}}$ can be substituted with other user-defined metrics, such as energy consumption or memory usage. 

\vspace*{-3mm}
\subsubsection{Pruning during supernet fine-tuning}
\label{sec:early_pruning}
\par Assume we have $D$ configurable architectural dimensions, such as the number of blocks, number of neurons, or number of heads per block. For each architectural dimension, we have $C$ different choices. Denote $p^d_c$ as the probability of the $c^{th}$ choice in the $d^{th}$ architectural dimension. The architectural entropy is defined as:
\begin{align}
H = - \frac{1}{D} \sum_{d=1}^D \sum_{c=1}^C p^d_{c} \log p^d_{c}.
\end{align}
When $H < \epsilon$, indicating convergence to a single sub-architecture, we prune all redundant branches, retaining only the optimal sub-architecture for continued fine-tuning.

\vspace*{-3mm}
\subsubsection{Knowledge Distillation}
\par After the pruning step and the supernet has converged to an optimal sub-network (when $H < \epsilon$), we continue fine-tuning it via a knowledge distillation approach, in which the largest subnet (the original model) acts as the teacher and the optimal sub-network serves as the student, thereby boosting the sub-network's accuracy. 

    



\vspace*{-4mm}
\subsection{Supernet design}

\subsubsection{Width dimensions}
\label{subsubsect:width}
\par
\begin{figure*}[!htb]
    \centering
    \includegraphics[width=0.9\textwidth]{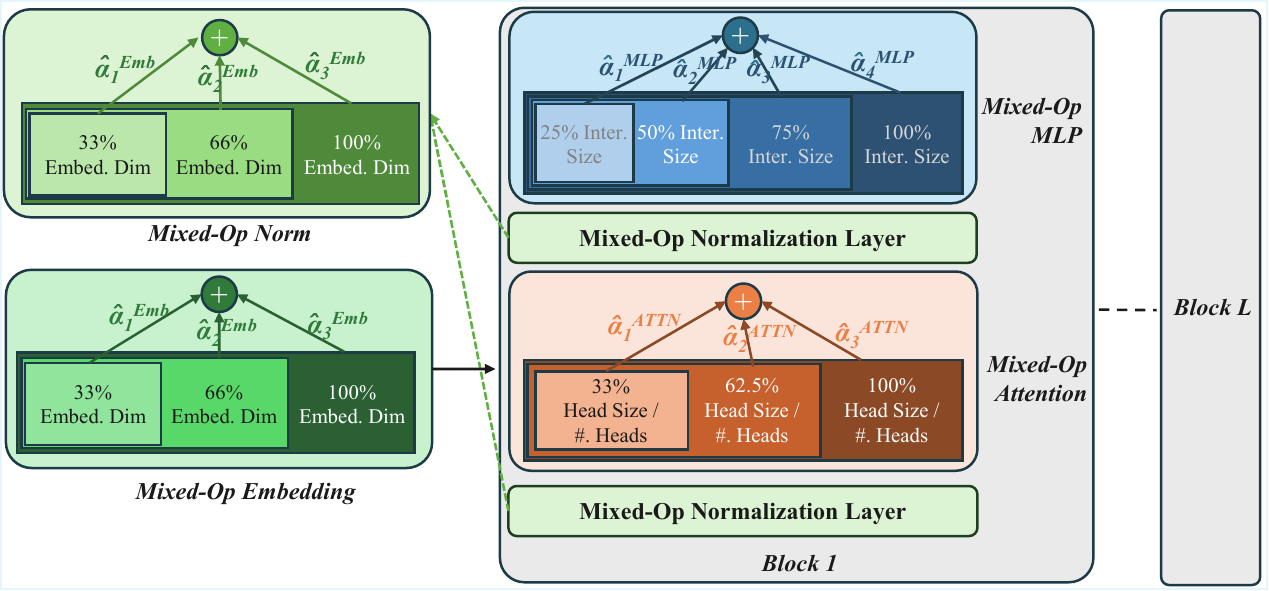}
    \vspace{-2mm}
    \caption{An overview of mixed-operation supernet design for width dimensions.}
    \label{fig:supernet_design}
    \vspace{-5mm}
\end{figure*}

 To differentiate between the width and depth aspects of the architecture, we denote $\alpha$ and $\beta$ as the respective parameters for these dimensions. Consequently, the overall architectural parameters are $\zeta = \{\alpha, \beta\}$. The width aspects encompass elements like the hidden size, count of attention heads, head size, and intermediate size, while the depth aspect pertains to the count of transformer blocks. Figure \ref{fig:supernet_design} depicts our supernet architecture, emphasizing the width dimension. Specifically, we introduce a mixed-operation weight for every embedding layer, normalization layer, and linear layer. This mixed weight is formed by a straightforward linear combination of possible sizes for the layer’s original weight, with contributions weighted by the sampled mixing coefficients $\hat{\alpha}$.
\par Drawing inspiration from DrNAS \cite{chen2021drnas}, we model the mixing weights $\hat{\alpha}$, which encode the relative importance or selection probability of each width option, as random variables drawn from a Dirichlet distribution parameterized by $\alpha$. We denote $\alpha^{IN}$ and $\alpha^{OUT}$ as the architectural parameters governing the input and output dimension selections for a given layer, respectively.
For a linear layer with base weight matrix $W_0$ of size $M \times N$, let $\hat{\alpha}i^{IN}$ and $\hat{\alpha}j^{OUT}$ represent the sampled mixing weights for the $i$-th input dimension choice $N^i$ and the $j$-th output dimension choice $M^j$, respectively. A sampling function $F_{sample}(W_0, N^i, M^j)$ extracts the top-left submatrix of $W_0$ consisting of the first $M^j$ rows and $N^i$ columns, then zero-pads it to restore the original $M \times N$ shape. As in TangleNAS \cite{sukthanker2024weight}, the mixed-operation weight $W_{mixed}$ for the layer is calculated accordingly.

\vspace*{-2mm}

\vspace{-2mm}
\begin{align} \label{eq:4}
W_\text{mixed} = \sum_{i} \sum_{j} \hat{\alpha}_{i}^\text{IN} \hat{\alpha}_{j}^\text{OUT} F_\text{sample}(W_{0}, N^{i}, M^{j})  \\
\text{with } \hat{\alpha}_{i}^\text{IN} \sim Dir(\alpha^{\text{IN}}_{i}),  \hat{\alpha}_{j}^\text{OUT} \sim Dir(\alpha^{\text{OUT}}_{j}) \notag
\end{align}  
\vspace*{-5mm}
\par The iterations in the loops of Equation \ref{eq:4} are computationally independent. Moreover, each iteration in the above equation requires a different padding size $(N-N^i, M-M^j)$. Therefore, the computation of the above equation requires non-uniform padding operations. However, Pytorch does not support parallelism for such operations and requires a loop over each $i$ and $j$ to run the slicing and padding operations. Consequently, naively computing the above equation on GPU with Pytorch, as in TangleNAS \cite{sukthanker2024weight} is expensive. Further details on this issue are discussed in Section \ref{sec:software}. 

\vspace*{-4mm}
\subsubsection{Depth dimension}
\label{subsubsect:depth_dimension}
 \par
\begin{figure}[!htb]
\centering
\begin{minipage}[c]{0.55\linewidth}
    \centering
    \subfigure[The mixed-op design]{\label{fig:depth_mixedop}\includegraphics[width=\linewidth]{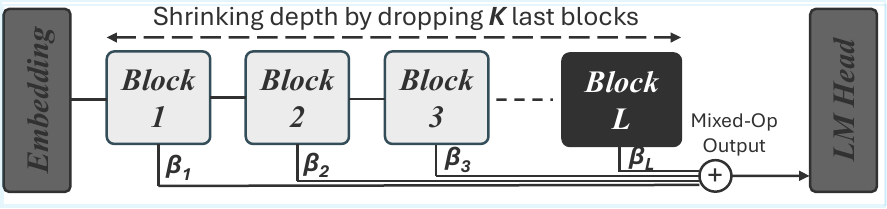}}
    \subfigure[Our design]{\label{fig:depth_reinmax}\includegraphics[width=\linewidth]{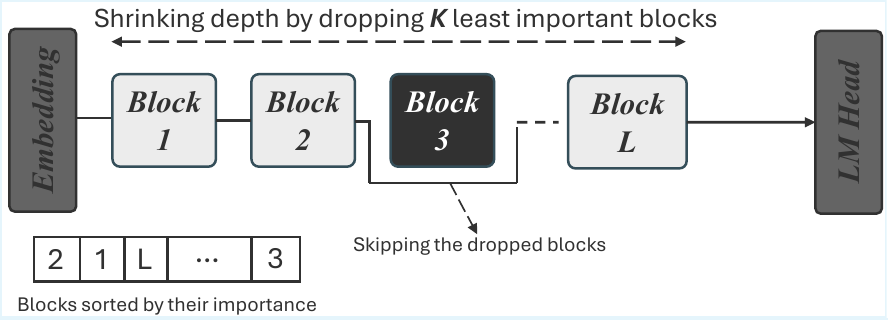}}
\end{minipage}%
\hfill
\begin{minipage}[c]{0.44\linewidth}
    \centering
    {\label{fig:your_label_c}\includegraphics[width=\linewidth]{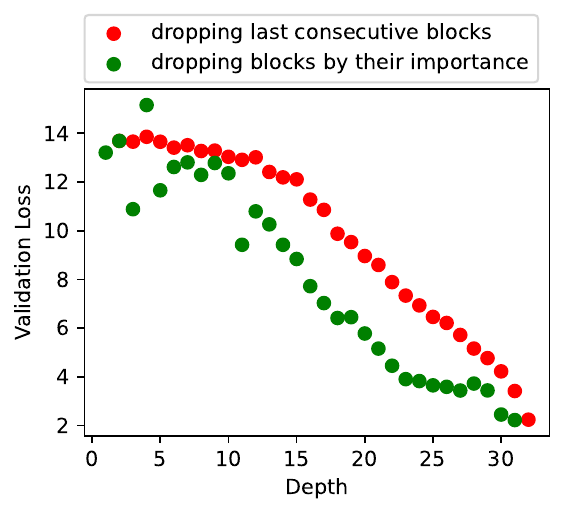}}
\end{minipage}
\vspace*{-2mm}
\caption{Left: The difference between the mixed-op design (a) and our proposed design (b) for pruning depth dimension. Right: Validation loss of compressed sub-network from Llama-3.1-8B model by dropping blocks with two schemes, namely dropping last consecutive blocks and dropping blocks by their importance.}
\label{fig:depth_design}
\vspace{-1mm}
\end{figure}

  Unlike the width dimensions, applying the mixed-operation approach to the depth dimension presents several challenges. Figure \ref{fig:depth_mixedop} illustrates the mixed-operation design for the depth dimension. This mixed-operation design requires to forward all $L$ blocks sequentially \cite{sukthanker2024weight}, and thus, it imposes strict constraints on depth reduction. Particularly, the final mixed-operation output for the block sequence is computed as a weighted sum of the outputs, with weights governed by the architectural parameters $\beta$. By optimizing $\beta$, when $\beta_{i} \to 1$, the mixed output is primarily influenced by the output of the $i^{th}$ block, effectively equivalent to sequentially forward all blocks up to the $i^{th}$ block and excluding all subsequent blocks.
 \par In a pre-trained large language model (LLM), transformer blocks have different levels of importance, and removing more critical blocks can significantly impair model performance compared to removing less essential ones. The block importance metric, first introduced by \cite{minitron2024}, quantifies a block’s sensitivity by measuring the cosine similarity between its input and output. We adopt this metric to assess block importance in our study. As illustrated on the right side of Figure \ref{fig:depth_design}, pruning blocks based on their importance results in a lower validation loss for the compressed sub-network compared to simply removing the final consecutive blocks. This finding aligns with \cite{sreenivas2024llm}, who examined the validation loss of the Llama-3 8B model when dropping 16 consecutive final blocks versus 16 non-consecutive blocks selected by importance. They found that dropping consecutive layers caused a significantly larger increase in validation loss compared to dropping non-consecutive layers.
 
 \par We propose an importance-aware depth pruning method to compress the depth dimension of large language models as in Algorithm \ref{alg:cap2}. Specifically, we model $\beta$ as parameters of a categorical distribution $Cat(\beta)$. In each training iteration, we sample the number of retained blocks $\hat{L} \sim Cat(\beta)$. To account for block importance, we maintain an array of block indices sorted by the importance of the corresponding blocks. For each sampled $\hat{L}$, we perform the forward pass using only the top $\hat{L}$ most important blocks, bypassing the rest. When $\beta_i \to 1$, it indicates that we retain the $i$ most important blocks and discard the remaining $L-i$ least important ones. 
 To enable gradient updates for $\beta$, we use a differentiable sampling technique, specifically the ReinMax method \cite{liu2023bridging}, a state-of-the-art gradient estimation technique for categorical sampling that provides enhanced robustness, as opposed to the Gumbel-Softmax trick described in \cite{jang2016categorical}. Figure \ref{fig:depth_reinmax} depicts our proposed design for the depth dimension.


    

\vspace*{-2mm}

    

\begin{algorithm}[!htb]
\caption{Importance-Aware Probabilistic Depth Pruning}
\label{alg:cap2}
\begin{algorithmic}[1]
\REQUIRE Pre-trained LLM with $N$ transformer blocks, learnable concentration parameters $\beta \in \mathbb{R}^{N}$, block importance scores $I \in \mathbb{R}^{N}$
\STATE $sorted\_indices \gets \text{argsort}(I)$ \COMMENT{Indices of blocks sorted from most to least important}
\FOR{each training iteration}
    \STATE $\hat{L} \sim Cat(\beta)$ 
    \STATE $kept\_indices \gets sorted\_indices[1 : \hat{L}]$ 
    \STATE Sort $kept\_indices$ in ascending order 
    \STATE $input \gets$ initial input to the first block
    \FOR{$i = 1$ \textbf{to} $\hat{L}$}
        \IF{$i \in kept\_indices$}
            \STATE $output \gets \text{block}_{i}(input)$ 
        \ELSE
            \STATE $output \gets input$ 
        \ENDIF
        \STATE $input \gets output$ 
    \ENDFOR
    \STATE Compute loss and backpropagate (gradients flow to $\beta$ via ReinMax reparameterization)
\ENDFOR
\end{algorithmic}
\end{algorithm}

\vspace{-1.1cm}

\subsubsection{Weight Quantization}
\par Fully fine-tuning the supernet described above is memory-intensive and impractical for deployment on a single NVIDIA A100 GPU. To mitigate this, we utilize LoRA techniques to reduce memory usage. Specifically, we incorporate LoRA adapters into all linear layers of the attention and MLP blocks. Let $Q_p(\cdot)$ represent the quantization function that quantizes an input matrix to precision $p$, with $A$ and $B$ denoting the matrices of the LoRA adapters. The standard approach to integrating LoRA with quantization involves quantizing the sum of the original weight and the LoRA adapter, as described in \cite{bondarenkolow,jeon2024l4q}. The quantized weight is formalized in Equation \ref{eq:6}.
\begin{align} \label{eq:6}
\tilde{W}^{} = Q_{p}(W_{0}  + B A)
\end{align} 
\par This design can eliminate mixed-precision computation overhead during the inference phase \cite{xu2024qa}. Specifically, we employ a unique pair of LoRA matrices A and B for each distinct weight quantization precision, rather than using a single pair for all precisions. This method addresses discrepancies in precision ranges among varying precisions, which could otherwise lead to instability in the supernet fine-tuning process. Let $P$ denote the set of weight precision. The mixed-op LoRA matrix for a precision $p$ is denoted as $BA_{p, mixed}$. The quantized mixed-op weight $\tilde{W}_{mixed}$ can be calculated as follows.
\begin{align} \label{eq:7}
\tilde{W}^{}_\text{mixed} =  \sum_{p\in P} \alpha^\text{weight}_{p}  * Q_{p}(W_\text{mixed} + BA_\text{p, mixed})
\end{align} 
Denote $\alpha^{weight}_{p}$ is the probability that the weight matrix $W$ being quantized to precision $p$. One advantage of our approach is its compatibility with various quantization-aware training (QAT) techniques \cite{jeon2024l4q,bondarenkolow}, enabling the integration of fine-tuning and QAT to obtain the most effective architectures, we defer this direction for future research. In this paper, we use Straight-Through Estimator~\cite{bengio2013estimating} to enable gradient flow through quantization operations.
\par Note that the QA-LoRA methods \cite{xu2024qa} are incompatible with our above design. Specifically, QA-LoRA alters the output shape of the $Q_{p}(.)$ function to $[M, \frac{N}{\text{group\_size}}]$ under low-precision settings, whereas it retains the $[M, N]$ shape for 16-bit precision, causing shape mismatches in the sum aggregation of Equation \ref{eq:7}.
\par To enhance processing speed through better GPU parallelism, we present a software enhancement in Section \ref{sec:software}. Regarding initialization, techniques for post-training quantization can be employed, including GPTQ \cite{frantar2023optq} or OmniQuant \cite{shaoomniquant}, among others. 

\vspace{-3mm}
\subsubsection{Activation Quantization}
We further explore activation quantization with a similar design. Particularly, the mixed activation $X_{mixed}$ of different precisions can be defined as follows:
\begin{align} \label{eq:8}
X_\text{mixed} =  \sum_{p\in P} \alpha^\text{activation}_{p}  Q_{p}(X)
\end{align}

\vspace{-5mm}
\subsection{Software Improvement: Vectorizing calculation of mixed-op weights}\label{sec:software}
\par
\begin{figure}[t]
\centering
\begin{minipage}[t]{0.52\linewidth}
    \centering
    \includegraphics[width=\linewidth]{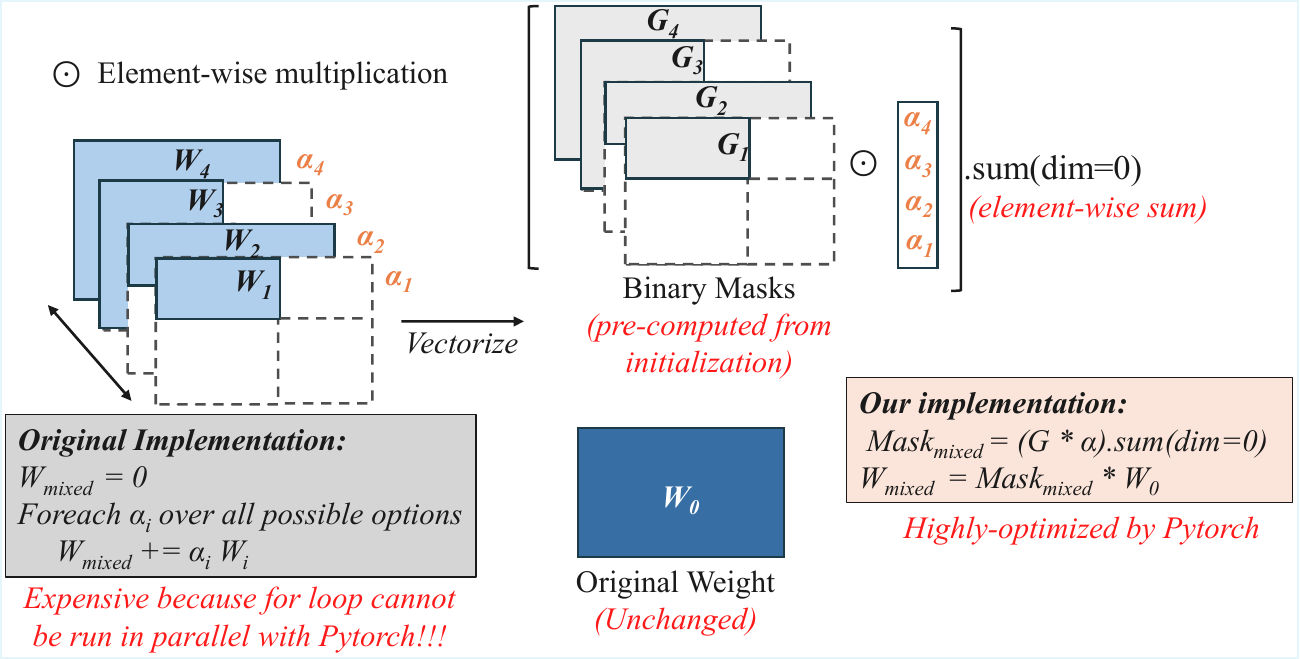}
\end{minipage}%
\hfill
\begin{minipage}[t]{0.45\linewidth}
    \centering
    \includegraphics[width=\linewidth]{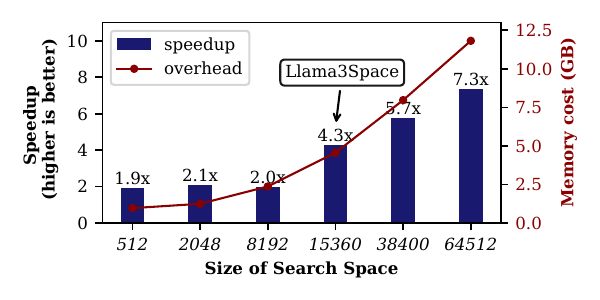}
\end{minipage}
\vspace{-2mm}
\caption{(Left) Software Implementation for mixed-op weight computation. (Right) Speedup and overhead memory cost of our proposed methods compared to the original weight-entanglement implementation \cite{sukthanker2024weight} with different numbers of candidate networks (sizes of search space) running on a A100 GPU with 80GB memory.}
\label{fig:software_speedup}
\vspace{-5mm}
\end{figure}

 The weight-entanglement method requires multiple nested loops to calculate combined weights and low-rank adaptations (LoRA) for each linear layer, which greatly slows down fine-tuning. Specifically, Equation \ref{eq:4} for the aggregated weight matrix is computationally expensive due to loops over indices $i$ and $j$, plus zero-padding that demands dynamic memory handling in each iteration. The left side of Figure \ref{fig:software_speedup} illustrates the original implementation and the implementation of our proposed software improvement for computing mixed weight \cite{sukthanker2024weight}.

\par To overcome this problem, we develop a method to vectorize the mixed weight calculation. Let $M_i$ and $N_j$ be the input and output feature sizes linked to architectural parameters $\alpha_i^\text{IN}$ and $\alpha_j^\text{OUT}$. We use a binary mask matrix $G_{i,j}$ (size $N \times M$) to indicate pairings between the $i$-th input and $j$-th output dimensions, where each element $g_{n,m}$ is:
\begin{align}
  g_{n,m} = \begin{cases}  \notag
1 & \text{if } n < N_{i}  \text{ and } m<M_{j},\notag  \\
0 & \text{otherwise}. \notag
\end{cases}
\end{align}
To fix the loop inefficiencies, we rewrite Equation \ref{eq:4} as:
\begin{align} \label{eq:5}
W_\text{mixed} = W_{0} \odot  \sum_{i}\sum_{j} G_{i,j}  \alpha_{i,j}
\end{align}
Here, $\odot$ is element-wise multiplication and $\alpha_{i,j} = \alpha_i^{IN}  \alpha_j^{OUT}$. The masks $G_{i,j}$ can be precomputed at model startup and reused across blocks. The sum involves multiplying a 3D tensor of all $G$ masks by the scalar vector $\boldsymbol{\alpha}$, then summing element-wise to create a 2D probabilistic mask that adjusts the base weights $W_0$. This vectorized version avoids loops, boosting fine-tuning speed. 

\par Although Equations \ref{eq:4} and \ref{eq:5} possess comparable complexity, our revised formulation supports parallel computation through broadcasting and element-wise operations. By precomputing the binary mask $G$ only once at initialization, it avoids the repeated slicing and padding required for calculating $W_{mixed}$ across layers in every training iteration. We additionally apply this updated approach to the consolidated LoRA matrices presented in Equation \ref{eq:7}. While our technique incurs a minor overhead for retaining the binary $G$ matrix, approximately 3.2 GB out of the 80GB available on an NVIDIA A100 GPU for the Llama-3.1-8B model using the \textit{Llama3Space} search space (outlined in Section \ref{sec:space}), this remains acceptable. Furthermore, the overhead is determined exclusively by the search space dimensions and is unaffected by size of training input prompts. In empirical tests with the \textit{Llama3Space} search space, this software enhancement provides up to 4.3$\times$ improved training throughput over the baseline weight-entanglement method from \cite{sukthanker2024weight}. This optimization can be adopted for any transformer-based architectures involving weight-entangled search spaces. 
 \par We evaluate the performance advantages of our software optimization strategy by benchmarking it against the standard weight-entanglement technique described in \cite{sukthanker2024weight}. This assessment includes determining the average runtime per training sample per iteration, using a batch size of 8, for both strategies across different search space scales, which are determined by the quantity of candidate networks. To vary the search space size, we adapt the \textit{Llama3Space} search space through modifications to architectural parameters, including the number of heads, embedding dimension, intermediate size, and head size. As illustrated in the right side of Figure \ref{fig:software_speedup}, the speedup results demonstrate the clear superiority of our software-driven enhancement compared to the baseline weight-entanglement method. Our approach delivers a 4.3$\times$ reduction in training time for the \textit{Llama3Space} search space, at the expense of an additional 3.2GB in memory usage. This enhancement stems from the fact that larger search spaces amplify the sequential loop iterations in Equation \ref{eq:4}, leading to increased runtime burdens in the original setup. In contrast, our vectorized method performs these operations concurrently, making it more resilient to expansions in search space size. Moreover, the speedup intensifies further as the number of candidate options rises, though this is accompanied by a corresponding increase in overhead memory. Therefore, when applying our method, it is crucial to consider the trade-off between memory costs and training acceleration.

\vspace{-0.3cm}
\section{Experiments}

\vspace{-0.1cm}
\label{Sect:Experiments}
\subsection{Baselines}
\vspace{-0.1cm}
\label{subsect:baselines}
\par We evaluate and compare our proposed approach with \textit{subnet-selection} \cite{sukthanker2024large} and LoNAS \cite{munoz2024lonas}, the state-of-the-art NAS approach for LLM compression. In addition, for a fair comparision, we also apply feature reordering based on their importance and sandwich training procedure when adapting LoNAS.
\par  Besides that, directly applying random search in search phase as in the original LoNAS paper \cite{munoz2024lonas} can lead to the dominance of small architectures in the result by the skewness in architectural distribution mentioned in \cite{sukthanker2024large}. Therefore, in the search phase of LoNAS, we apply the same grid-based partition strategy as in\cite{sukthanker2024large} by splitting the search space into $k$ even parts, then randomly searching the best architecture from each partition.

\vspace{-0.3cm}
\subsection{Dataset}
\vspace{-1mm}
\par \textbf{Calibration and Fine-tuning Dataset:} To ensure a fair comparison with other state-of-the-art methods, we adopt the same experimental setup as described in \cite{sukthanker2024large}. Specifically, we use the same calibration dataset from \cite{sukthanker2024large} for computing important score.  We utilize the Alpaca dataset \cite{alpaca} with approximately 52,000 instructions. The fine-tuning dataset is divided into two separate train and validation datasets during search phase of our method. After the supernet converges to a single optimal sub-architecture, we further fine-tune this optimal sub-architecture using the whole dataset.

\par \textbf{Evaluation Dataset:} We evaluate and compare our method with other baselines with 7 diverse common-reasoning tasks, namely BoolQ \cite{clark2019boolq}, PIQA \cite{Bisk2020}, HellaSwag (HS) \cite{zellers2019hellaswag}, WinoGrande (WG) \cite{sakaguchi2020winogrande} and ARC \cite{allenai:arc} (ARC-easy (ARC\_E) and ARC-challenge (ARC\_C)), and MMLU \cite{hendryckstest2021}. Moreover, similar to experimental settings in \cite{sukthanker2024large}, we perform 5-shot evaluation for WinoGrande, 10-shot for HellaSwag, 25-shot for ARC-challenge, 5-shot evaluation for MMLU, and 0-shot evaluation for remaining tasks.

\vspace{-0.2cm}
\subsection{Latency Profiling}
\vspace{-1mm}
To evaluate and measure the inference latency of the Llama3 model across different weight-activation quantization setups, we utilize several state-of-the-art kernels tailored to each configuration. Specifically, for W4A16 and W8A16 quantization, we adopt the Marlin kernel \cite{frantar2025marlin}, which achieves top-tier performance in these scenarios. For alternative quantization configurations, we apply the method introduced in ABQ-LLM \cite{zeng2025abq}, enabling support for arbitrary quantization in linear layers. This profiling and data gathering process is conducted on an A100 GPU equipped with 80GB of memory.

\vspace{-0.2cm}
\subsection{Search Space}\label{sec:space}
\par We use Llama-3.1-8B as the foundation LLM, however, it is remarkable that our method is adaptable to other foundation models. For fair baseline comparisons, we assess our method against baselines within a search space called  \textit{Llama3Space}. For our quantization extension, we employ a distinct search space named  \textit{QLlama3Space}. Notably, we adopt a group size of 128 for quantization options in \textit{QLlama3Space}, which requires us to exclude certain choices for hidden size, number of heads, and head dimensions that are not divisible by 128. Details of both search spaces are provided in Alpaca\ref{tab:search_space}. For the sake of simplification, we denote \textbf{\textit{our-no-quant}} and \textbf{\textit{our-quant}} as our proposed method run with \textit{Llama3Space} and \textit{QLlama3Space} search space, respectively.
\vspace{-0.3cm}
\begin{table}[h]\scriptsize
\centering
\caption{Possible configurations of search space for Llama-3.1 8B}
\label{tab:search_space}
\begin{tabular}{|c|c|c|}
\hline
\textbf{Dimensions} & \textbf{Llama3Space} & \textbf{QLlama3Space} \\
\hline
Hidden Dim. & $\{2^i | i \in [5, 12]\}$ & $\{2^i | i \in [7, 12]\}$ \\
Number of Heads & $\{8, 16, 32\}$  & $\{ 16, 32\}$ \\
 Head Dim. & $\{8, 16, 32, 64, 128\}$ & $\{32, 64, 128\}$ \\
Intermediate Dim. & $\{4096 * i| i\in [1.0, 2.0, 3.0, 3.5]\}$ & $\{4096 * i| i\in [1.0, 2.0, 3.0, 3.5]\}$ \\
Number of Blocks & $\{1,...,32\}$ & $\{1,...,32\}$ \\
Weight Bitwidth & & $\{2, 4, 8\}$ \\
Activation Bitwidth & & $\{2, 4, 8, 16\}$ \\
\hline
\textbf{\# of Candidates} & $8 \times 3 \times 5 \times 4 \times 32=15360$ & $6 \times 2 \times 3 \times 4 \times 3 ^ {32} \times 4 ^ {32} \times 32$  \\
\hline
\end{tabular}

\vspace{-4mm}
\end{table}

\vspace{-1mm}


\vspace{-0.5cm}
\subsection{Experimental Details}
\vspace{-2mm}
\par For our method, \textit{subnet-selection}, and LoNAS, LoRA is set with a rank of 32, alpha of 16, and a dropout rate of 0.05 during fine-tuning. We also pre-process the pre-trained LLM by reordering neurons and attention heads by their important scores as in \cite{sukthanker2024large}. To compute the block importance, we follows the method as mentioned in \cite{minitron2024}. Unlike \cite{sukthanker2024large}, which applies LoRA solely to embedding and attention layers, we follow LoNAS \cite{munoz2024lonas} by applying LoRA to both attention and MLP layers.  Training settings for LoRA adapters are similar as settings in \cite{sukthanker2024large}. 


\par In terms of the quantization approach, we utilize group-wise quantization with a group size of 128. Furthermore, we initialize the quantizers in our proposed supernet using OmniQuant \cite{shaoomniquant}. For quantizing models suggested by the baselines (\textit{subnet-selection} and LoNAS), we similarly employ the OmniQuant method. Notably, other post-training quantization techniques can also be applied to initialize our supernet.

\vspace{-0.3cm}

\subsection{Comparison to baselines}
\vspace{-0.1cm}
\subsubsection{Without Quantization}
\begin{figure}[t]
\centering
\begin{minipage}[t]{0.45\linewidth}
    \centering
    \includegraphics[width=\linewidth]{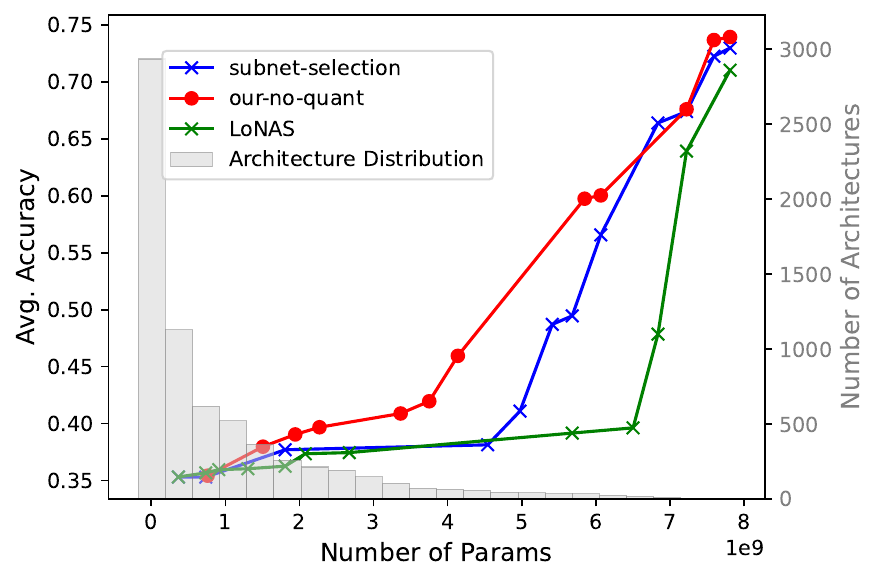}
    \vspace{-2mm}
    \caption{Pareto fronts of our proposed method compared to \textit{subnet-selection} and LoNAS. The second y-axis is the architectural distribution across different ranges of model size for \textit{Llama3Space} search space.}
    \label{fig:paretofronts_1}
\end{minipage}%
\hfill
\begin{minipage}[t]{0.52\linewidth}
    \centering
    \includegraphics[width=\linewidth]{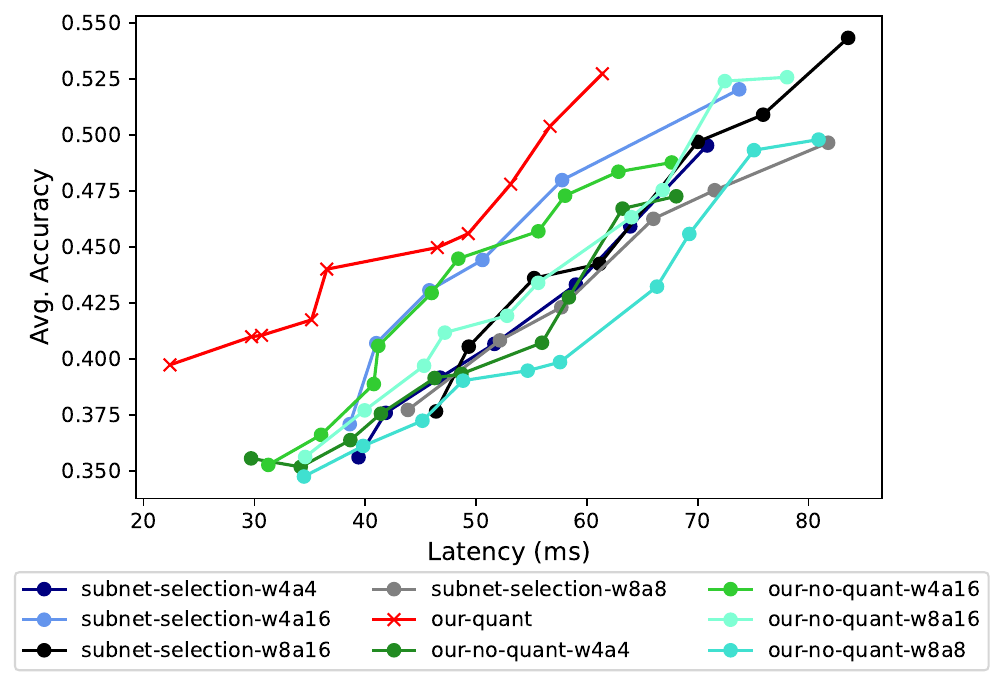}
    \vspace{-2mm}
    \caption{Pareto fronts of \textbf{\textit{our-quant}} method compared to models compressed by \textit{subnet-selection} and \textbf{\textit{our-no-quant}}, with 4 different quantization configurations, namely W4A4, W4A16, W8A8, and W8A16}\label{fig:paretofronts_3}
\end{minipage}
\vspace{-5mm}
\end{figure}

Table \ref{tab:sim} presents a comparison of our proposed method against baseline approaches across common reasoning tasks, evaluated over four distinct parameter ranges. Notably, our method \textbf{\textit{our-no-quant}} surpasses state-of-the-art baselines in both average accuracy and inference latency. To illustrate the advantages of our approach, Figure \ref{fig:paretofronts_1} shows Pareto fronts comparing our method to baselines, alongside the distribution of sub-architectures across various model size ranges. The distribution reveals a skew, with smaller models being far more prevalent than larger ones. This observation align with observations in previous work \cite{sukthanker2024large}. Specifically, in the 6–8 billion parameter range, only a few architectures are present. When compared to two baselines, our method demonstrates superior performance across the 2–6 billion parameter range. For larger models, the architectures suggested by our method align closely with those from the \textit{subnet-selection} method, primarily due to the limited number of available architectures in this range, allowing both methods to identify optimal configurations. However, in the 2–6 billion parameter range, our method consistently outperforms all baselines, indicating its ability to identify superior compression configurations. Besides that, \textit{subnet-selection} also outperforms LoNAS in this range of model size. Therefore, for the experiment with quantization (Section \ref{subsect:quantize}), we apply quantization to \textit{subnet-selection} and use it as the baseline for comparison. For small models ($<$ 2 billion parameters), the performance gap between our method and the baselines is minimal, as highly compressed architectures lose significant information, resulting in reduced accuracy across all methods. 

\begin{table*}[t] \scriptsize
\centering
\caption{A comparison between baselines with our proposed methods for 7 common reasoning tasks, with 4 different ranges of model size, namely 2-3 billion (2-3B), 3-4 billion (3-4B), 4-5 billions (4-5B), and 5-6 billions (5-6B) parameters. Remarkably, LoNAS* are improved as mentioned in Section \ref{subsect:baselines}.}\label{tab:sim}

\begin{tabular}{l c c c c c  c c c c c}
\hline
Method & Params &  \makecell[c]{Latency\\(ms)} & \makecell[c]  ARC\_E & ARC\_C& BoolQ & WG &	HS	& MMLU	& PIQA & \makecell[c]{Acc.\\(\%)} \\
\hline
     \multicolumn{11}{c}{2B-3B} \\
\hline
 \textit{subnet-selection} & 2.54B & 70.95 & 25.66 & \textbf{26.62} & 49.88	&  38.50  & 26.38 & \textbf{25.72} & 53.76 & 35.22 \\
LoNAS* & 2.67B & 69.84 & 24.45 &  25.77 & 48.86 & \textbf{61.87} & 26.05 & 22.95 & 52.23 & 37.45   \\
 \textbf{\textit{ours-no-quant}} & 2.27B & \textbf{43.14} & \textbf{34.18} & 23.98 & \textbf{60.83} & 50.67 & \textbf{27.43} & 25.38 & \textbf{55.28} & \textbf{39.68}   \\
\hline
     \multicolumn{11}{c}{3B-4B} \\
\hline
 \textit{subnet-selection} & 3.28B & 81.31 & 28.20 & 25.77 &	48.78 &	37.95 &	27.15 &	24.49 & 	54.03 &	35.19 \\
 LoNAS & 3.28B &  81.31  & 24.49 & \textbf{27.47} & 49.72 & 38.69  &	26.58 &	24.65 &	51.41 & 34.72 \\
  \textbf{\textit{ours-no-quant}} & 3.37B & \textbf{40.90}  & \textbf{38.13} & 23.21 & \textbf{60.55} & \textbf{52.72} & \textbf{29.21} &	\textbf{25.86} & \textbf{56.53} & \textbf{40.89}  \\
\hline

    \multicolumn{11}{c}{4B-5B} \\
\hline
 \textit{subnet-selection} & 4.32B & 57.45  & 31.99 &	26.88 &	37.86 &	49.49 & 31.59 & 25.07 & 57.73 &  37.22 \\
   LoNAS* & 4.00B & 73.72 & 25.25 & 26.11 & 46.96 & 56.67  & 26.82 &	23.74 & 52.84 & 36.91   \\
   \textbf{\textit{ours-no-quant}} & 4.14B & \textbf{53.93} & \textbf{41.33} &  \textbf{29.79} & \textbf{62.69} & \textbf{56.27} & \textbf{41.98} & \textbf{26.90} & \textbf{62.68}  & \textbf{45.93} \\
\hline
    \multicolumn{11}{c}{5B-6B} \\
\hline

\textit{subnet-selection} & 6.06B & \textbf{87.38} & \textbf{61.41} &	37.12 &	70.43 &	\textbf{65.59} &	59.96 &	30.98 &	70.46 &	 56.56 \\
 LoNAS* & 6.84B & 99.00 & 42.59 & 31.83 & 57.77 & 63.12 &	 51.00 & 24.26 & 64.42 & 47.86   \\
  \textbf{\textit{ours-no-quant}} & 6.06B  & \textbf{87.38} & 59.18 &	\textbf{42.32} & \textbf{72.54}	& 65.51	& \textbf{60.77} & \textbf{49.07}	& \textbf{70.82}  &  \textbf{60.03}\\
  \hline
\end{tabular}

\vspace{-0.3cm}
\end{table*}  

    
    

\vspace{-0.1cm}
\subsubsection{With Quantization}
\label{subsect:quantize}
In this experiment, we investigate the \textit{QLlama3Space} search space, which supports a wide range of quantization configurations. For a fair comparison, we apply post-training quantization to the architectures identified by the \textit{subnet-selection} and \textbf{\textit{our-no-quant}} methods using four quantization settings: W4A4, W8A8, W8A16, and W4A16. These quantized models are then benchmarked against those produced by our proposed joint optimization approach, denoted \textbf{\textit{our-quant}}. We exclude the LoNAS baseline from this analysis, as \textit{subnet-selection} consistently outperformed it in earlier experiments.
Figure \ref{fig:paretofronts_3} presents the Pareto fronts of compressed architectures discovered by our-quant, compared against those from \textit{subnet-selection} and \textbf{\textit{our-no-quant}} under the various quantization settings. The results clearly show that jointly optimizing both architectural configurations and quantization precisions enables our method to achieve superior efficiency–accuracy trade-offs compared to sequential pipelines (architecture search followed by quantization). For example, at the same average accuracy of 40\% across reasoning tasks, models compressed with our method achieve up to $1.4\times$ faster inference than those from competing baselines. Alternatively, at a fixed inference latency of 30 ms, our models reach approximately 41\% average accuracy on reasoning tasks, outperforming other baselines by roughly 6\%.

\vspace{-0.3cm}
\section{Conclusions and Future Work}
\vspace{-2mm}
\label{Sect:Conclusion}
\par We present an effective compression approach for large language models (LLMs) using differential neural architecture search. Notably, our non-quantized method, \textbf{\textit{ours-no-quant}}, outperforms state-of-the-art approaches for non-quantized configurations, while our quantized version, \textbf{\textit{ours-quant}}, which jointly optimizes both compression and quantization, surpasses sequential compression followed by quantization. This suggests that jointly optimizing structural architecture and quantization yields more efficient compression configurations compared to treating them as separate problems. 

\vspace{-0.3cm}
\bibliographystyle{splncs04}
\bibliography{references}
\end{document}